# An Effective Method for Fingerprint Classification

Monowar Hussain Bhuyan, Sarat Saharia, and Dhruba Kr Bhattacharyya
Department of Computer Science & Engineering, Tezpur University, India.

**Abstract** *This paper presents an effective method for fingerprint classification using data mining approach. Initially, it generates a numeric code sequence for each fingerprint image based on the ridge flow patterns. Then for each class, a seed is selected by using a frequent itemsets generation technique. These seeds are subsequently used for clustering the fingerprint images. The proposed method was tested and evaluated in terms of several real-life datasets and a significant improvement in reducing the misclassification errors has been noticed in comparison to its other counterparts.*

## 1. Introduction

Fingerprints have been used as the most popular biometric authentication and verification measure because of their high acceptability, immutability and uniqueness [18]. Immutability refers to the persistence of the fingerprints over time whereas uniqueness is related to the individuality of ridge details across the whole fingerprint image. Fingerprint classification is an important step in any fingerprint identification system because it significantly reduces the time taken in identification of fingerprints specially where the accuracy and speed are critical. Classification allows an input fingerprint to be matched against only by a subset of a database and is critical in speeding-up fingerprint identification. Conversely, classification is not enough to identify a fingerprint; it is useful in deciding when two fingerprints do not match. To reduce the search and space complexity, a systematic partitioning of the database into different classes is highly essential.

Key to the task of classification is the feature extraction. The effectiveness of feature extraction depends on the quality of the images, representation of the image data, the image processing models, and the evaluation of the extracted features. At the first stage of the fingerprint classification process, the image is only represented as a matrix of grey scale intensity values. Feature extraction is a process through which geometric primitives within images are isolated in order to describe the image structure, i.e. to extract important image information and to suppress redundant information that are not useful for classification and identification processes. Thus fingerprint features and their relationships provide a symbolic description of a fingerprint image.

In this paper, we present an effective method to classify the fingerprint images using a partitional clustering technique. For classification of the fingerprint images, it initially determines a compact representation (i.e. a 32-dimensional numeric sequence or pattern) for each fingerprint image based on the ridge flow pattern by exploiting the freemen chain code approach [28]. Then it exploits the *apriori* [3] algorithm to identify the seeds (i.e. the maximal frequent itemsets) over the meta-base for the cluster formation. Once the seeds are determined, a *k*-means like clustering algorithm was applied to determine the classes. The robust *apriori* based seed selection and the simplified *k*-means logic makes the classification method more attractive in view of the following points: (*i*) overcomes the local minimum problem and (*ii*) linear cluster formation time.

The rest of the paper is organized as follows: section 2 reports a brief survey. In section 3, the background of the work is discussed. Section 4 presents the classification method based on partitioning approach. It also reports the proof of correctness of the proposed classification method substantiated with some experimental results. Finally, in section 5, the concluding remarks are given.

## 2. Related Works

Based on our survey related to fingerprint classification, it has been observed that most of the existing works are aimed to classify the fingerprint database based on the minutiae sets, singular points and other techniques [15, 25]. In this section, some of these are reported in brief.

- Masayoshi Kamijo's Classification approach [12]: It is an ANN based approach, where a neural network for the classification of fingerprint images is constructed, which can classify the complicated fingerprint images. It uses a two-step learning method to train the four-layered neural network which has one sub-network for each category. It carries out the principal component analysis (PCA) with respect to the unit values of the second hidden layer and also studies the fingerprint classification state represented by the internal state of the network. Consequently, the method confirms that the fingerprint patterns are roughly classified into each category in the second hidden layer and the effectiveness of the two-step learning process.

- However, in case of larger data sets this method can be found to give limited results.
- Karu and Jain's Classification approach [13]: In this approach, it initially finds the ridge direction at each pixel of an input fingerprint image. Then the algorithm extracts global features such that singular points (cores and deltas) in the fingerprint image and performs the classification based on the number and locations of the detected singular points. Here, the singular point(s) detection is an iterative regularization process until the valid singular points are detected. If the images are of poor quality, the algorithm classifies those images as unknown types based on some threshold values. However, the algorithm can detect the labeled images with high quality.
- Ballan and Ayhan Sakarya's Classification Technique [2]: Here, a fast, automated, feature-based technique for classifying fingerprints is presented. The technique extracts the singular points (deltas and cores) in the fingerprints based on the directional histograms. It finds the directional images by checking the orientations of individual pixels, computes directional histograms using overlapping blocks in the directional image, and classifies the fingerprint into the Wirbel classes (whorl and twin loop) or the Lasso classes (arch, tented arch, right loop, or left loop). The complexity of the technique is the order of the number of pixels in the fingerprint image. However, it takes much time for classification.
- Jain, Prabhakar and Hong's Multi-channel Classification approach [10]: This method can be found to be more accurate while classifying the fingerprint images as compared to its previous counterparts. Here, the fingerprint images are classified into five categories: whorl, right loop, left loop, arch, and tented arch. The algorithm uses a novel representation (FingerCode) and is based on a two-stage classifier to make a classification. The two-stage classifier uses a k-nearest neighbor classifier in its first stage and a set of neural network classifiers in its second stage to classify a feature vector into one of the five fingerprint classes. This algorithm suffers from the requirement that the region of interest be correctly located, requiring the accurate detection of center point in the fingerprint image. Otherwise, the algorithm can be found to be very effective.
- Cho, Kim, Bae et al's Classification approach [5]: They described a new fingerprint classification algorithm that uses only the information related to the core points. The algorithm detects core point(s) candidates roughly from the directional image and analyzes the near area of each core candidate. In this core analysis, false core points made by noise are eliminated and the type and the orientation of core point(s) are extracted for the classification step. Using this information, classification was performed. However, it can be found to be very difficult to eliminate the false singular point(s) which has been used for class decision. It demands for more sophisticated methods to eliminate those false core points towards a noise-tolerant classification system.
- Yao, Marcialis, Pontil, et al's Classification approach [29]: Here, a new fingerprint classification algorithm is reported based on the two machine learning approaches: support vector machines (SVMs), and recursive neural networks (RNNs). RNNs are trained on a structured representation of the fingerprint image. They are also used to extract a set of distributed features which can be integrated in the SVMs. SVMs are combined with a new error correcting coding scheme, which unlike previous systems, can also exploit information contained in ambiguous fingerprint images.
- Tan, Bhanu and Lin's Fingerprint Classification [24]: Here, a fingerprint classification approach was proposed based on a novel feature-learning algorithm. They used Genetic Programming (GP) based approach, which learns to discover composite operators and features that are evolved from combinations of primitive image processing operations. They developed an approach to learn the composite operators based on primitive features automatically. It can be found to be useful in the extraction of some useful unconventional features, which are beyond the imagination of humans. They also defined the primitive operators as very fundamental and easy to compute. Then, primitive operators are separated into computation operators and feature generation operators. Features are computed wherever feature generation operators are used. This classification method can be found to be effective over quality fingerprint images.
- Shah and Sastry's Classification approach [22]: It classifies the fingerprint images into one of the five classes: arch, left loop, right loop, whorl, and tented arch. It can be found to be useful over low-dimensional feature vector obtained from the output of a feedback based line detector. Their line detector was a co-operative dynamic system that gives oriented lines and preserves multiple orientations at points where differently oriented lines meet. Also the feature extraction process of their method was based on characterizing the distribution of orientations around the fingerprint. Three types of classifiers used here: support vector machines, nearest-neighbor classifier, and neural network classifier. The line detector works on binary images only.
- Park and Park's classification approach [21]: Here, a new approach for fingerprint classification is reported based on discrete Fourier transform (DFT) and nonlinear discriminant analysis. Utilizing the DFT and directional filters, a reliable and efficient directional image is constructed from each fingerprint image, and then nonlinear discriminant analysis is applied to the constructed directional images, reducing the dimension drastically and extracting the discriminant features. The method explores the capability of DFT and directional filtering in dealing with low-quality images and the effectiveness of nonlinear feature extraction method in fingerprint classification.
- Ji, Zhang Yi's SVM-based Fingerprint Classification [11]: Here, a classification method of fingerprint images using orientation field and support vector machine is reported. It estimates the orientation field through pixel gradient values, and then calculates the percentages of the directional block classes. These percentages are combined as a four dimensional feature vector, by which the trained hierarchical classifier classifies the fingerprints into one of the six classes. The supervised training rule is adopted to train the hierarchical classifier with five one-against-all SVMs. Fingerprints can be classified into one of the six classes using this trained classifier.
- Wei, Yonghui, et al's Classification approach [26]: In this approach, a new fingerprint classification algorithm is introduced that uses some curve features of the ridgelines. The algorithm basically exploits the total direction changes of the ridgelines during classification. However, the sampling of the ridgelines still can be found to be time consuming.

## 2.1. Discussions

Based on our overall survey following observations are made:
- Several methods have been proposed in the past couple of years to address the fingerprint classification issues. Most of these methods classify the images based on the ridges, local feature (i.e. minutiae) and global features (i.e. singular points).
- Model based approaches based on the global features (singular points) of the fingerprints have been found more effective in classifying the fingerprints into different known classes.
- Structure-based approaches based on the estimated orientation field in a fingerprint image can be found capable to classify the images into one of the five classes. The role of the estimated orientation field for fingerprint classification is generic. However, if the images are of poor quality then the orientation field estimation could not be done properly. Also, in such case difficulties encountered during extraction of other features like minutiae, finger code, Poincare index for singular points detection etc.

In this work, prior to classification a numeric meta-base was created by using the freeman chain code approach over the original grey-scale fingerprint images. Then a robust frequent itemsets generation technique was used to select the seeds for the cluster formation. Subsequently, a linear time unsupervised classification technique was used based on the seeds. Next, we provide the background of our classification technique.

## 3. Background of the work

Real-time image quality assessment can greatly improve the accuracy of identification system. The good quality images require minor pre-processing and enhancement. Conversely, low quality images require major pre-processing and enhancement.

## 3.1. Fingerprint Image Pre-processing

Chen *et al*. [4] used fingerprint quality indices in both the frequency domain and spatial domain to predict image enhancement, feature extraction and matching performance. L. Hong [9] proposed the enhancement of the fingerprint image using filtering technique. Lim *et al*. [16] computed the local orientation certainty level using the ratio of the maximum and minimum eigen values of the gradient covariance matrix and the orientation quality using the orientation flow. The main steps involved in the pre-processing include: (*a*) enhancement (*b*) binarization (*c*) segmentation, and (*d*) thinning. Next, each of these steps is described in brief.

- Image Enhancement: The input fingerprint image is pre-processed on both the spatial and frequency domain. In the spatial domain, histogram equalization technique was applied for better distribution of the pixel values over the image to enhance the perceptional information. In the frequency domain, the image was divided into small processing blocks (32×32 pixels) and the Fast Fourier transform (FFT) [27] was applied in the following way –

$$F(u, v) = \sum_{x=0}^{M-1}\sum_{y=0}^{N-1} f(x,y) \times \exp\left\{-j2\pi \times \left(\frac{ux}{M}+\frac{vy}{N}\right)\right\} \quad (1)$$

Where $u = 0, 1, 2,...., 31$ and $v = 0, 1, 2,..., 31$. In order to enhance a specific block by its dominant frequencies, the FFT of the block was multiplied by its magnitude for a number of times. Here, the magnitude of the original FFT = $abs(F(u,v)) = |F(u,v)|$.

$$g(x, y) = F^{-1}\{F(u,v) \times |F(u,v)|^k\} \quad (2)$$

where $F^{-1}(F(u, v))$ is done by:

$$f(x,y) = \frac{1}{MN}\sum_{x=0}^{M-1}\sum_{y=0}^{N-1} F(u,v) \times \exp\left\{j2\pi \times \left(\frac{ux}{M}+\frac{vy}{N}\right)\right\} \quad (3)$$

for $x = 0, 1, 2, ..., 31$ and $y = 0, 1, 2, ..., 31$.
The value of $k$ in equation (2) is an experimentally determined constant, however based on our experimentation a better result was found for $k =0.55$. It has been observed that for a higher value of $k$ the appearance of the ridges improves, filling up small holes in the ridges; however, with a too high $k$ results in false joining of ridges. So with an appropriate selection of $k$ value, the ridges and the overall appearance of the image can be improved, which is useful for proper feature extraction and classification.
- Image Binarization: In this step, an 8-bit grey level fingerprint image is transformed into a 1-bit image with 1-value for ridges and 0-value for furrows. In [31], an optimized approach can be found for binarization. However, in our work, an enhanced binarization method was used which is based on the adaptive binarization approach [27]. In this method, the pixel value is transformed to 1 if the value is larger than the mean intensity value of the current block (16×16 pixels); otherwise, it is set to zero.

- Image Segmentation: The objective of the fingerprint segmentation is to extract the region of interest (ROI) which contains the desired fingerprint impression. Fingerprint image segmentation highly influences the performances of Automatic Fingerprint Identification System (AFIS). In our work, the gradient based fingerprint segmentation approach [1] was used and the segmentation results were found satisfactory even for the low quality images.
- Image Thinning: This step aims to eliminate the redundant pixels of ridges till the ridges are just one pixel wide. Here, an iterative, parallel thinning algorithm was used for thinning the binarized fingerprint image [14]. In each scan of the full fingerprint image, the algorithm marks down redundant pixels in each small image window (3×3 pixels) and finally removes all those marked pixels after several scans. Also, in our experimentation, the advancement of each trace step still has large computational complexity although unlike other thinning algorithms, it does not require the pixel by pixel movement.

## 3.2. Fingerprint Feature Extraction and Numeric Meta-base Creation

Fingerprint image offers a rich source of information for classification and matching of fingerprints. However, for a given raw image, automatic extraction of the fingerprint features is an extremely difficult task, specially, when the fingerprint acquisitions are noisy (which is common). The effectiveness of a fingerprint classification system is based on the extracted features from the fingerprint image. There are mainly two types of features [19] that are useful in fingerprint identification systems: (*i*) Local ridges and furrows details (minutiae), which have different characteristics of different fingerprints, and (*ii*) Global pattern configurations which form the special patterns of ridges and furrows in the central region of the fingerprints. The first type of features carries information about the individuality of the fingerprints, whereas the second type carries information about the class of fingerprints. For effective recognition, the extracted features should be invariant to the translation and rotation of the fingerprint images. Generally, global features can be derived from the orientation fields and from the shapes of global ridges. The orientation field of a fingerprint consists of the ridge tendencies in local neighbourhoods and forms an abstraction of the local ridge structures.

In our approach, an orientation image was created from the binarized fingerprint image using the gradient based approach [1]. In order to create the image, the core point was determined based on the density of the directional flow of the fingerprint image and it is basically for the four classes of fingerprint images i.e. tented arch, left loop, right loop and whorl. The binarized image, orientation image and the core point of the image are shown in figure 1(a), (b), and (c).

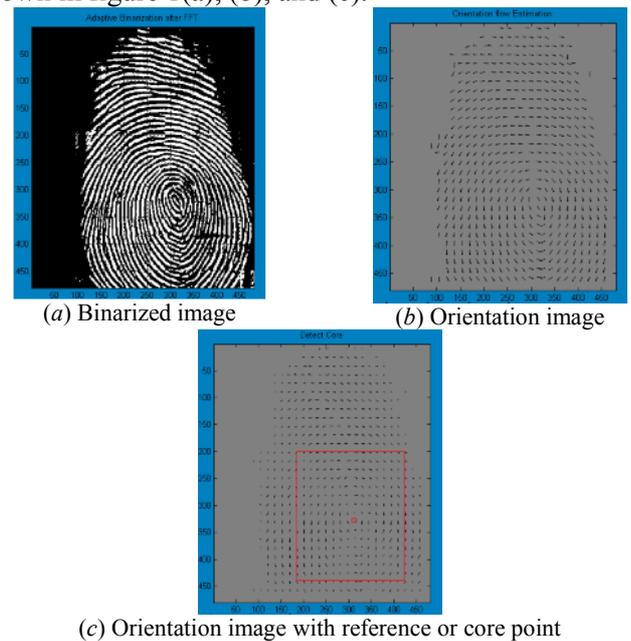

(*a*) Binarized image  (*b*) Orientation image

(*c*) Orientation image with reference or core point
Figure1. Fingerprint image

### 3.2.1. Numeric Meta-base Creation

To create the meta-base, $T_{mxn}$, where $m$ and $n$ are the number of rows (i.e. the number of images in the databases) and the number of dimensions respectively, each pre-processed image was divided into 3×3 grids (shown in *figure 2*). Now, the initial grid cell i.e. $R_p$ is chosen which is basically the *core* or *reference* point in case of those four classes other than *arch*, however, in case of *arch*, it is the starting point of that ridge for which the inter-ridge distance can be found to be least. Then following this initial grid cell, the successive control points were chosen along the ridge flow of the image and accordingly numeric codes (0-7) are assigned depending on the regulation pattern and its directionality. The idea behind the orientation flow codes is that each pixel of an image can have almost 8-neighbours and thus the direction from a given pixel can be specified by a unique number from 0 to 7 [28] as shown in figure 2. Each of these numbers represents one of the eight possible directions from the given pixel along the ridge flow of the given image and figures 3(a) and (b) show the orientation image without and with grid space. This is continued for $n$-successive points. The parameters used in generating $T$ are reported in Table 1. Table 2 reports the corresponding values of ridge (for the successive control points) and directionality for each numeric code. In our experimental study, we found better result for the number of control points i.e. $n = 32$ for representing an image.

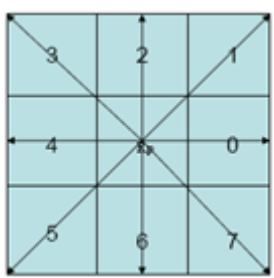

Figure 2. 3×3 grid and 8-connected codes

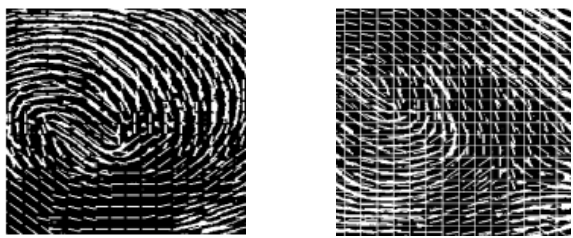

(*a*) Orientation flow image  (*b*) Orientation flow image with 3×3 grid

Figure 3. Orientation flow image

Table 1. Code generation parameters

| Grad of $x$-$\delta_x$ | Grad of $y$-$\delta_y$ | Numeric Code | Directionality, $Z = (4*(\delta_x + 2)+(\delta_y + 2))$ |
|---|---|---|---|
| 0 | 1 | 0 | 11 |
| -1 | 1 | 1 | 7 |
| -1 | 0 | 2 | 6 |
| -1 | -1 | 3 | 5 |
| 0 | -1 | 4 | 9 |
| 1 | -1 | 5 | 13 |
| 1 | 0 | 6 | 14 |
| 1 | 1 | 7 | 15 |

Table 2. Parameters for assigning numeric codes

| Ridge value, $RPV_i$ | Directionality, $Z_i$ | Numeric code |
|---|---|---|
| 1 | 11 | 0 |
| 1 | 7 | 1 |
| 1 | 6 | 2 |
| 1 | 5 | 3 |
| 1 | 9 | 4 |
| 1 | 13 | 5 |
| 1 | 14 | 6 |
| 1 | 15 | 7 |

An example ridge flow pattern matrix is shown in *figure 4*.

$$T_{mxn} = \begin{bmatrix} 0\ 0\ 1\ 2\ 3\ 3\ \ldots\ldots\ 6\ 7\ 7 \\ 0\ 4\ 5\ 5\ 7\ 7\ \ldots\ldots\ 4\ 3\ 2 \\ 0\ 4\ 4\ 5\ 7\ 6\ \ldots\ldots\ 3\ 3\ 2 \\ \vdots \\ 0\ 0\ 1\ 1\ 3\ 4\ 5\ \ldots\ldots\ 5\ 6\ 7 \end{bmatrix}$$

Figure 4. Ridge flow pattern matrix

The logic for creation of $T_{mxn}$ is shown in the procedure *NMC*. It takes image $I_i$ as input and invokes the procedure *GetCore* to find the core point over $I_i$. *GetCore* works based on [27]. It uses the variable $RPV_i$ to represent the ridge pixel value of the image and $Z_i$ to refer the condition instance of the $Z$ values from *Table 1*. The variable $C_p$ represents the control points.

**Procedure *NMC*( )**
*Input*: Fingerprint image $I_i$ (i = 1, 2, 3, .......,m)
*Output*: Ridge orientation flow codes matrix $T_{mxn}$
  1. for i = 1 to m do
  2. read the image $I_i$;
  3. call GetCore() to find the core point $C_i(x, y)$ ;
  4. for each core point $C_i(x, y)$ do
     4.1 for $C_p$ = 1 to n do
        [a] if $RPV_i$ = 1 and satisfies $Z_i$ then
             update $T_{mxn}$ with corresponding
             code (lookup Table 2)
     4.2 next $C_p$;
  5. next i.

### 3.3. Clustering Using Partitioning Approach

The partitioning clustering approach partitions the database of *n*-objects into *k*-number of predefined clusters where each partition represents a cluster and $k \leq n$. It also satisfies two conditions (*i*) each cluster must have at least one object, and (*ii*) each object must belong to exactly one cluster. Next, for the sake of understanding the *k*-means clustering technique is reported in brief.

#### 3.3.1. *K*-means Clustering

The *k*-means clustering [17] is a frequently used clustering algorithm that minimizes an objective function; this algorithm assumes that the number of clusters in which the data set will be classified is known.

The *k*-means algorithm can be divided into two phases: the *initialization* phase and the *iteration* phase. In the initialization phase, the algorithm randomly assigns the object into *k*-clusters, whereas in the iteration phase, the algorithm computes the distance between each non selected objects and each cluster representative and assigns the object to its nearest cluster. The algorithm works in linear time; however, it has the local minimum problem. So, during the initialization phase, if special care is taken in selecting the *k* initial seeds, it can overcome the local minimum problem.

#### 3.3.2. Effectiveness of Proximity Measures

In the unsupervised classification method, the proximity or distance measure plays an important role. To have high accuracy in the classification, an appropriate selection of the distance measure is highly essential. In this section, the experimental results are reported on the suitability of the various dissimilarity measures with *k*-means based classification approach over high dimensional numeric database.

We have experimented with four distance measures: *Euclidean distance*, *City block distance*, *cosine distance* and *correlation measure* with *k*-means clustering over FVC datasets and results are reported in *Table 3*. To evaluate the effectiveness of the distance measure, *misclassification error* [20] was computed using equation (4).

$$M_E = \frac{1}{N}\sum_{i=1}^{k}|(cluster(D_i) - (cluster(D_i')))| \quad (4)$$

where $D_i$ is the total number of cluster objects before clustering, $D_i'$ is the total number of cluster objects after clustering, *k* is the total number of clusters and *N* is the total number of points in the datasets. It can be observed from the table that the performance of the *k*-means based classification using *Euclidian distance* is better than the other distance measures.

Table 3: Simple *k*-means cluster analysis

| Dataset Size | Euclidean Distance | City-block Distance | Cosine Distance | Correlation |
|---|---|---|---|---|
| 32 | 0.125 | 0.312 | 0.125 | 0.125 |
| 300 | 0.060 | 0.126 | 0.126 | 0.060 |
| 800 | 0.150 | 0.132 | 0.175 | 0.210 |
| 1710 | 0.260 | 0.260 | 0.260 | 0.220 |
| 2300 | 0.204 | 0.248 | 0.207 | 0.250 |
| 2900 | 0.031 | 0.230 | 0.202 | 0.250 |

Next, the proposed unsupervised classification method based on the modified *k*-means clustering using *Euclidean distance* is discussed.

## 4. FPCLU: Fingerprint Clustering Method

The proposed FPCLU aims to classify any input fingerprint database into any of the five categories as reported in *figure 5*.

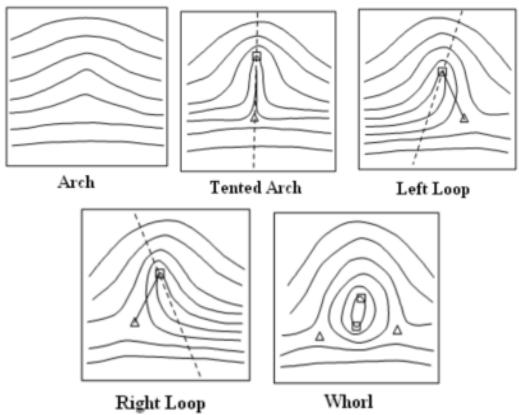

Figure 5. A prototype of each fingerprint class

FPCLU classifies the images using an unsupervised approach over the meta-base. Initially, it selects the probable seeds for each class using the frequent itemsets generation technique, i.e. *apriori*(). Once the seeds i.e. the maximal frequent itemsets over the meta-base are selected, *k*-means like clustering is applied for the formation of the clusters. Next, definitions and lemma which provide the theoretical basis of the classification method are introduced.

- Definition 1: A pattern tuple $Tt_i$ of a ridge flow patterns matrix *T* is defined as an *n*-dimensional tuple where the value of each element in the tuple is $0 \leq Lt_i \leq 7$.
- Definition 2: Two ridge flow pattern tuples are defined as coherent or similar if no. of similar occurrences between the tuples $\geq \delta_i$, where $\delta_i$ is a user defined threshold.
- Definition 3: Seed of a cluster can be defined as one of the distinct maximal frequent ridge flow pattern tuples of *T* with respect to $\alpha$ (where $\alpha$ is a user defined parameter).
- Definition 4: A ridge flow pattern tuple $Tt_i$ can be defined as noise, iff (*i*) $Tt_i \notin C_i$ (where $C_i$ is the instance of the set of clusters) and (*ii*) $Tt_i$ is not maximally frequent.
- Lemma 1: A seed will lead to a distinct class.
  Proof: Let $S_i$'s (*i* = 1 to 5) be the seeds and let $S_i$ corresponds to $C_1$ and $C_2$, i.e. two distinct classes. Now, as per *definition 3*, a seed is a distinct maximal frequent itemsets (i.e. ridge flow pattern tuple) and as per *definition 2*, a non-selected pattern tuple $Tt_j$ will be a member of a class represented by a seed say $S_i$, if they are found to be coherent. Hence, $S_i$ will not lead to $C_1$ and $C_2$ both; it contradicts.

### 4.1. Seed Selection Using *Apriori()*

In case of partitioning based clustering, generally the clustering accuracy largely depends on the initial seeds. A major disadvantage of *k*-means clustering is that it has no guarantee for an optimal solution. To overcome this problem, the algorithm needs to be executed multiple times in order to find better clustering results. However, there is no guarantee for the global optimum solution. In our approach, to overcome it the initial seeds are selected by using a robust frequent itemsets generation technique i.e. *apriori*() [23]. *Apriori* helps to identify the maximal frequent itemsets over the numeric meta-base, which were subsequently filtered out for the identification of five distinct pattern (seeds) w.r.t. to a user defined threshold $\alpha$. These seeds were finally used by the procedure FPCLU for classifying the fingerprint images. The seed selection process basically follows the following steps:

- Find the highly correlated patterns of frequent itemsets using *apriori*(); (*ii*) Take intersection among them; if the number of common items $\geq \alpha$, then accept it as a seed for cluster expansion; (*iii*) repeat step (*ii*) for the next class (until seeds for all the five classes are identified) for subsequent seeds. An example of this seed selection process is reported in *Table 4*. In our experiment, the best result (least misclassification error) was obtained with threshold value $\alpha \geq 3$. Next, the classification method is reported.

Table 4 reports the frequent itemsets generated by *apriori* and the corresponding seed for each class determined based on the intersection of the frequent itemsets. In *column 1 & 2* of the table, the transaction ID and the maximal frequent itemsets are shown. In *column 3* the seeds for all these classes w.r.t. $\alpha \geq 3$ are shown which are determined based on the intersection of the corresponding maximal frequent itemsets and finally in *column 4* the unique transaction ID that corresponds to the seeds are shown.

Table 4. An example of seed selection for $\alpha \geq 3$ based on Apriori() results

| TID | Frequent itemsets generated by Apriori | Seeds | Unique TID |
|---|---|---|---|
| T0003, T0013, T0020, T0056, T0088, T0112 | 3, 6, 9, 13, 18, 24 | 3, 6, 9, 18 | T0013 |
| T0003, T0013, T0020, T0061, T0093, T0112 | 3, 6, 9, 14, 18, 27 | | |
| T0123, T0154, T0188, T0221, T0257, T0276 | 6, 9, 14, 19, 27, 29 | 6, 9, 14, 27 | T0257 |
| T0123, T0154, T0188, T0235, T0257, T0269 | 6, 9, 14, 23, 27, 31 | | |
| T0285, T0301, T0357, T0403, T0498 | 4, 10, 12, 19, 29, 31 | 4, 10, 19, 29 | T0498 |
| T0285, T0301, T0318, T0403, T0498 | 4, 10, 14, 19, 25, 29 | | |
| T0504, T0539, T0566, T0589, T0601, T0632 | 2, 7, 11, 25, 26, 32 | 7, 11, 32 | T0601 |
| T0539, T0566, T0578, T0593, T0601, T0632 | 7, 11, 19, 25, 29, 32 | | |
| T0539, T0566, T0580, T0599, T0601, T0632 | 7, 11, 13, 26, 29, 32 | | |
| T0544, T0563, T0645, T0691, T0767, T0779 | 6, 9, 13, 21, 24, 32 | 6, 9, 21, 32 | T0767 |
| T0544, T0563, T0605, T0672, T0767, T0779 | 6, 9, 12, 21, 28, 32 | | |

## 4.2 Proposed Classification Method: FPCLU

This routine takes $T_{mxn}$ and $\alpha$ as input and generates five distinct classes i.e. $C_1, C_2, \ldots, C_5$. Steps of the algorithm are:

1. read $T_{mxn}$
2. call Apriori() to find maximal frequent itemsets $L_i$'s
3. select five seeds $S_i$'s from the $L_i$'s w.r.t $\alpha$.
4. [a] take a non-selected object and find its dissimilarities with $S_i$'s based on Euclidean distance;
   [b] assign class ID to the object of that $S_i$ found to be similar;
5. repeat step 4 until no more objects to be classified.

### 4.2.1. Complexity Analysis of FPCLU

Considering the meta-base creation as a pre-processing task, the proposed classification method basically works in two phases. Phase-I is dedicated in the seed selection using apriori() which takes $O(n^2)$ complexity. In phase-II, the seeds determined in the previous phase are used for cluster formation which works in linear time. Thus the complexity due to phase-II is $O(kn)$. Hence, the total complexity is $O(n^2) + O(kn)$.

### 4.2.2. Experimental Results

- Environment used: The experiment was carried out on an hp xw8600 workstation with Intel Xeon Processor (3.33GHz) with 4GB RAM. MATLAB 7.6 (R2008a) revised version in windows (64-bits) platform was used for the performance evaluation.
- Datasets used: Our algorithm was tested on the standard databases: FVC 2000 [6], FVC 2002 [7] and FVC 2004 [8]. These databases totally consist of 9600 fingerprint images from 330 different fingers, composed of various classes. However, only 2900 images were chosen for experimentation. Out of these, 300 images are *arch*, 200 are *tented arch*, 890 are *left loop*, 900 are *right loop*, and 610 are *whorl*.
- Clustering Effectiveness: To evaluate the performance of the proposed FPCLU, the equation for misclassification errors (Equation no. 4) was used for the various distance measures, and the results are reported in Table 5. From the table, it can be seen easily that FPCLU gives better result for Euclidean distance. Now, to evaluate its performance while comparing with the other counterparts, accuracy and false acceptance rate (FAR) were computed for the various sample sizes. The averaged result of the proposed method is reported in Table 6. From the table it can be seen that most of the better methods (in terms of accuracy) can identify successfully only four classes. However, our method can identify five classes successfully. However, the accuracy of FPCLU also depends on accurate identification of the reference or core point.

Table 5. Fingerprint classification error analysis using modified k-means algorithm

| Dataset Size | Euclidean Distance | City-block Distance | Cosine Distance | Correlation |
|---|---|---|---|---|
| 32 | 0.093 | 0.12 | 0.12 | 0.093 |
| 300 | 0.033 | 0.046 | 0.06 | 0.033 |
| 800 | 0.07 | 0.10 | 0.11 | 0.08 |
| 1710 | 0.031 | 0.058 | 0.039 | 0.029 |
| 2300 | 0.022 | 0.071 | 0.053 | 0.036 |
| 2900 | 0.017 | 0.050 | 0.051 | 0.033 |

An exhaustive analysis of the proposed method in terms of false acceptance rate (FAR) vs. classification accuracy was also carried out in light of FVC datasets. The results and its comparison with its other counterparts are reported in figure 6

The GUI for FPCLU based classification method was implemented using MATLAB. It can be found in http://agnigarh.tezu.ernet.in/~dkb/ is reported in figure 7

Table 6. A comparisons of the recent fingerprint classification algorithm accuracies with FPCLU algorithm

| FPCLU and other counterparts | Classes | Accuracy(%) |
|---|---|---|
| Wilson et al. [27] | 5 | 81.0 |
| Karu and Jain [7] | 5 | 85.4 |
| Jain et al. [27] | 5 | 90.0 |
| Hong and Jain [9] | 5 | 87.5 |
| Cappelli et al. [27] | 5 | 92.2 |
| Yao et al. [11] | 5 | 89.3 |
| Chang and Fan[27] | 5 | 94.8 |
| Mohamed and Nyongesa[27] | 5 | 92.4 |
| Zhang et al.[27] | 5 | 84.0 |
| Zhang et al. [27] | 5 | 84.0 |
| Yao et al.[27] | 5 | 90.0 |
| Wilson et al. [27] | 4 | 86.0 |
| Karu and Jain [27] | 4 | 91.4 |
| Jain et al. [27] | 4 | 94.8 |
| Hong and Jain[27] | 4 | 92.3 |
| Senoir [27] | 4 | 88.5 |
| Jain and Minut [27] | 4 | 91.3 |
| Yao et al.[27] | 4 | 94.7 |
| Mehran and Gheysari[27] | 4 | 99.02 |
| **FPCLU** | **5** | **98.3** |

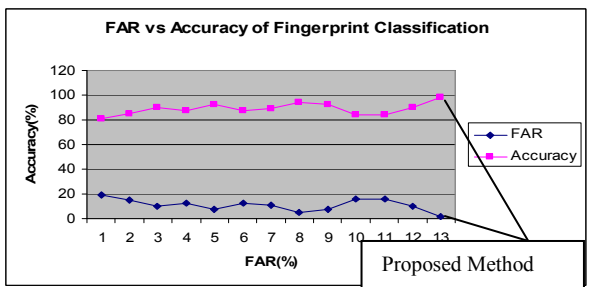

Figure 6. FAR vs. Accuracy of fingerprint classification

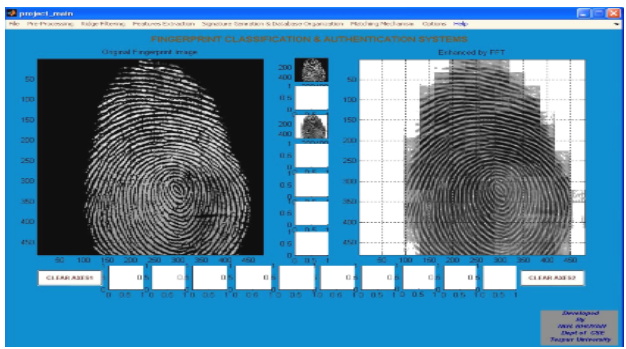

Figure 7. GUI for fingerprint classification and authentication

## 5. Concluding Remarks

A limited survey on some of the popular fingerprint classification methods has been carried out and reported in this paper. These methods basically operate on the minutiae sets and singular points. However, most of these methods have been found capable of identifying four or five classes with an accuracy level of (80-95) %. In this paper, an effective method for fingerprint classification with an accuracy level of 98% has been reported. The method exploits the robust *apriori* algorithm for the seed selection and the linear *k*-means like algorithm for the cluster formation based on those seeds. Further research is going on towards development of an effective search method over the large clustered fingerprint databases.

## Acknowledgement
This work is a part of a research project and authors are thankful to AICTE for funding the project.

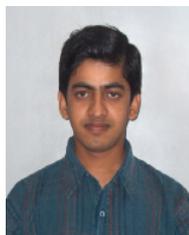

**Monowar Hussain Bhuyan** received his AMIETE degree in Computer Science & Engineering from The Institution of Electronics and Telecommunication Engineers (IETE) in 2007 and received his M.Tech. degree in Information Technology from Department of Computer Science & Engineering, Tezpur University in 2009. Now, he is pursuing his Ph.D degree in Computer Science & Engineering from the same University. He is a life member of IETE, India. His research areas include Biometric Authentication, Data Mining, and Network Security.

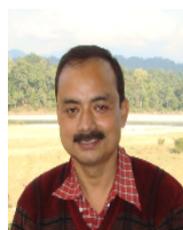

**Sarat Saharia** received Master of Computer Application (MCA) degree from Dibrugarh University, India in 1993. He joined Tezpur University, India as a Lecturer of Computer Science Department in 1995 and currently he is an Associate Professor in Computer Science and Engineering department of the same University. He is pursuing Ph.D. in the same department and expected to submit his thesis shortly. His research interest includes pattern recognition, optical character recognition and image processing.

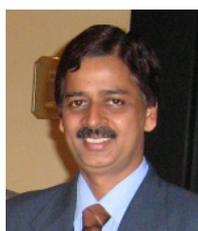

**Dhruba Bhattacharyya** did his PhD in Computer Science from Tezpur University in 1999. Presently he is serving as a Professor in the Computer Science & Engineering Department at Tezpur University. His research areas include Data Mining, Network Security and Content based Image Retrieval. Prof. Bhattacharyya has published more than 100 research papers in the leading Int'nl Journals and Conference Proceedings. Also, Dr Bhattacharyya written/edited 04 books. He is a Programme Committee/Advisory Body member of several Int'nl Conferences/Workshops.